\documentclass[runningheads]{llncs}
\usepackage[T1]{fontenc}
\usepackage{graphicx}
\usepackage{xcolor}
\usepackage{amsmath, amssymb}
\usepackage{svg}
\usepackage{hyperref}
\begin{document}
\title{From \textit{What} to \textit{Why}: Thought-Space Recommendation with Small Language Models}
\titlerunning{Thought-Space Recommendation with SLM}
\author{Prosenjit Biswas \and
Pervez Shaik\thanks{This work was carried out during an internship at Sony Research India.} \and
Abhinav Thorat \and
Ravi Kolla \and
Niranjan Pedanekar}
\authorrunning{Prosenjit Biswas et al.}
\institute{Sony Research India\\
\email{\{prosenjit.biswas,
shaik.pervez,
abhinav.thorat, 
ravi.kolla, 
niranjan.pedanekar\}@sony.com}}

\maketitle              % typeset the header of the contribution
\begin{abstract}
Large Language Models (LLMs) have advanced recommendation capabilities through enhanced reasoning, but pose significant challenges for real-world deployment due to high inference costs. Conversely, while Small Language Models (SLMs) offer an efficient alternative, their reasoning capabilities for recommendation remain underexplored. Existing systems often use natural language rationales merely as unsupervised descriptive text, failing to harness their full potential as learning signals. In this work our main idea is to create a common understanding of user and items across multiple domains called \textit{Thought Space} with SLMs instead of using LLMs' distilled knowledge. To that end we propose \textbf{PULSE} (\textbf{P}reference \textbf{U}nderstanding by \textbf{L}atent \textbf{S}emantic \textbf{E}mbeddings), a framework that treats SLM-generated rationales as director learning signals, supervising them with interaction histories to jointly model user actions (\textit{what}) and their semantic drivers (\textit{why}). Existing methods consider only interactions such as sequences and embeddings, whereas PULSE treats rationales as first-class signals, this novel design yields embeddings that are more robust and generalizable. Extensive experiments demonstrate that PULSE outperforms leading ID, Collaborative Filtering (CF), and LLM-based sequential recommendation models across multiple benchmark datasets. Furthermore, PULSE exhibits superior transferability in cross-domain recommendation and demonstrates strong performance on downstream tasks such as reasoning-oriented question answering. Our code is available \href{https://anonymous.4open.science/r/Thinking_PULSE-0FC5/README.md}{\underline{here}}.
\end{abstract}

\section{Introduction}
\label{sec:introduction}
Recommender systems form the backbone of modern digital platforms, enabling users to navigate vast catalogs in e-commerce, media, and social domains. The field has long relied on collaborative filtering and sequence modeling techniques to predict user preferences, but these approaches struggle in settings with sparse interactions, cold-start users, and cross-domain transfer. Recently, Large Language Models (LLMs) have opened new possibilities by bringing broad world knowledge, semantic reasoning, and generative abilities into recommendation.

The integration of LLMs with recommender systems has given rise to multiple emerging research directions. Prompt-based strategies reformulate user and item IDs into natural language prompts \cite{wu2022selective,li2023personalized}, enabling LLMs to perform zero-shot or few-shot recommendation. Embedding-fusion hybrids \cite{acharya2023llm} merge collaborative filtering IDs with LLM-derived semantics to build richer user and item representations. More recent reasoning-augmented models go beyond outputs, using LLMs to generate intermediate rationales or reasoning chains that explain why a user may prefer an item examples include R2Rec~\cite{zhao2025reason} and RDRec~\cite{wang2024rdrec}, which utilize interaction graphs or LLM-generated rationales into signals for downstream models. These works show that LLMs can enhance both accuracy and explainability in recommender systems.

However, the limitations of combining LLMs and recommender systems are equally evident. The dependency on billion-scale parameters makes LLM-based recommenders costly to train, slow to serve, and difficult to deploy in latency-sensitive environments. Even distillation with SLMs requires LLM inference, which leads to significant computational overhead. Moreover, reasoning, despite being a defining strength of LLMs remains underexplored; rationales are typically consumed without systematic supervision, leaving their potential as direct learning signals untapped.

This motivates growing interest in small language models (SLMs), which offer a practical balance of efficiency and representational power. Recent work \cite{xu2024slmrec}, \cite{wang2024can} shows that carefully distilled or optimized SLMs can match or even surpass LLM baselines in sequential recommendation. For instance, SLMRec finds that large model size adds little to the next-item prediction task, as distilled models achieve comparable accuracy with only 13\% of the parameters while running 6–8× faster. Similarly, Lite-LLM4Rec~\cite{wang2024rethinking} eliminates costly text generation by restructuring the model for direct candidate scoring, cutting inference latency by over 97\%. These results indicate that much of the brute-force capacity of LLMs is redundant for recommendation, while well-designed SLMs centered on task-specific reasoning can achieve both accuracy and efficiency.

LLM-based next-item prediction tasks rely heavily on reasoning. However, it remains largely unexplored whether smaller language models (SLMs) can leverage rationales as primary learning signals. Rationales capture why users engage, not just what they engage with, offering semantic and causal structure beyond interaction IDs. With proper alignment, rationale-driven representations can improve accuracy and cross-domain generalization, as reasoning patterns (e.g., “prefers organic products across categories” or “seeks long-term effectiveness over short-term trends”) transfer even when interaction data does not, making them well-suited for contrastive learning.

Prior work \cite{wu2021self}, \cite{qiu2022contrastive} has applied contrastive learning (CL) between multiple views of interactions, for example by masking sequences, perturbing graphs, or injecting noise into embeddings. To the best of our knowledge, existing works have treated interactions as the primary unit of contrast, typically defining positives as alternate views of the same user’s history and negatives as samples from other users or unobserved items. Since rationales offer deeper insight into user preferences, they present a compelling basis for contrastive learning beyond interaction-level signals. What remains unexplored is contrasting rationales’ embeddings, which capture natural language reasons for preferences and are explicitly optimized as part of the learning objective.

In this work, we take a step towards filling these gaps. We propose \textbf{PULSE} (\textbf{P}reference \textbf{U}nderstanding by \textbf{L}atent \textbf{S}emantic \textbf{E}mbeddings), a framework that builds an SLM based generalized Thought Space, that encodes user behaviour understanding, from inter and intra-domain contrastive objectives, and then supervises the rationales generated from the same SLM, for the current domain. We summarize the key contributions of this work as follows:
\begin{enumerate}
    \item \textbf{Thought Space for user modeling.} We introduce a novel embedding space where rationales and behaviors are contrastively aligned, enabling a more generalized yet discriminative representation of users.
    
    \item \textbf{Rationale-level contrastive learning.} Unlike prior work that applies contrastive objectives only on interactions (sequences, graphs, embeddings), we contrast rationales as supervised anchors: positive rationales from ground-truth interactions vs.\ negatives from other users/domains.
    
    \item \textbf{SLM-driven reasoning.} We show that small language models (Phi-4, 4B) can generate rationales that, when aligned in Thought space, rival or surpass LLM-based methods. Rationales are not consumed verbatim but refined as supervised training signals.
    
    \item \textbf{Empirical gains.} On sequential recommendation benchmark datasets, our method PULSE improves HR@1 by a range of 12-27\% over all baselines.
    
    \item \textbf{Generality.} Our approach transfers across domains, yielding $\sim$30\% higher HR@1 in cross-domain setups, and further boosts reasoning-heavy QA (HotpotQA), surpassing state-of-the-art F1 and EM scores.
\end{enumerate}

\section{Related Works}
\label{sec:related-works}
\textbf{LLMs in Recommendation.} LLMs have recently been adapted to recommendation, offering strong reasoning and representational power. Early works reformulated IDs as prompts to tackle cold-start and explainability challenges \cite{wu2022selective,li2023personalized}, while others enriched item profiles with natural language descriptions or multimodal summaries \cite{acharya2023llm,zhou2023mmrec}. More advanced approaches combine LLM reasoning with graph-based modeling \cite{wang2024llmrg} or collaborative filtering knowledge \cite{kim2024large,zhang2025collm}. These studies demonstrate the promise of LLMs but highlight challenges in scale, inference cost, and deployment, as noted in recent surveys \cite{shehmir2025llm4rec}.
 
\textbf{SLMs in Recommendation.}  SLMs provide a cost-effective alternative, achieving competitive performance at a fraction of the inference cost \cite{wu2024could}. Knowledge distillation and sequential recommendation have been explored \cite{xu2024slmrec,shridhar2023distilling}, yet SLMs are still largely treated as lightweight surrogates to LLMs. Their reasoning abilities remain underexplored, with no prior work supervising or refining rationales as part of training.
 
\textbf{Rationales in Recommendation.}  Rationales, natural language explanations for user–item interactions enhance transparency, personalization, and trust. Early works extracted salient reviews for explanations \cite{pan2022accurate}, while recent studies leverage LLM-generated reasoning for personalization, user profiling, and multimodal recommendation agents \cite{tsai2024leveraging,bismay2024reasoningrec,zhang2025reasonrec}. However, rationales are typically used \emph{post hoc}, without refinement or integration as a direct learning signal.
 
\textbf{Contrastive Learning (CL) in Recommendation.}  CL has been widely applied to improve robustness and representations, contrasting multiple views of user–item interactions via graph perturbations, embedding noise, or explanation-aware augmentations \cite{wu2021self,qiu2022contrastive,yu2022graph,cai2023lightgcl,wang2022explanation}. Yet, CL has never been applied to rationales themselves. Existing methods treat interactions as the sole object of augmentation, overlooking reasoning artifacts that encode intent and causality. Leveraging rationales as contrastive views may yield richer, semantically aligned embeddings and facilitate cross-domain generalization.

\section{Problem Formulation}
\label{sec:problem-formulation}
\begin{table}[t]
\centering
\small
\caption{Summary of key notations used in this work.}
\label{tab:notations}
\begin{tabular}{|l|p{10cm}|}
\hline
\textbf{Notation} & \textbf{Description} \\
\hline
$\mathbf{s}_u$ & Historical interaction sequence of user $u$ \\
\hline
$i_{t+1}$ & Ground-truth next item for user $u$ \\
\hline
$\mathcal{C}_u = \{c_1,\dots,c_{10}\}$ & Candidate items set: 9 negatives + 1 ground truth \\
\hline
$d_i$ & Textual description (metadata) of item $i$ \\
\hline
$\rho_{u,i}, v_{u,i}$ & Rating and review of user $u$ on item $i$ \\
\hline
$b_u = (\mathbf{s}_u, i_{t+1})$ & Behavioral text (history + ground truth) \\
\hline
$R_u^+$ & Positive rationale for user $u$ \\
\hline
$\{R_{v_j}^-\}$ & Negative rationales from other users $v_j$ \\
\hline
$E_1, E_2$ & Rationale encoder and behavior encoder, respectively \\
\hline
$\mathbf{z}_r, \mathbf{z}_h$ & Embeddings of rationale and behavior context \\
\hline
$\widehat{\mathcal{R}}_u$ & Candidate reasons generated via ToT for user $u$ \\
\hline
$S(\mathcal{R}_{i,j})$ & Cosine similarity score between reason $\mathcal{R}_{i,j}$ \& behavior embedding $b_u$ \\
\hline
$S_{\max}$ & Maximum similarity score across all candidate reasons \\
\hline
$\mathcal{R}_{S_{\max}}$ & Best reason (leaf) attaining the maximum similarity score $S_{\max}$ \\
\hline
$f_\phi$ & Fine-tuned SLM scoring head (parameters $\phi$ are trainable) \\
\hline
$p_\phi$ & Softmax probability over candidates predicted by the SLM \\
\hline
$\mathcal{L}_{\text{CL}}$, $\mathcal{L}_{\text{CE}}$ & Contrastive and Cross-Entropy losses respectively \\
\hline
\end{tabular}
\end{table}

%\subsection{Task: Sequential Recommendation} 
Let $\mathcal{U}$ and $\mathcal{I}$ denote the sets of users and items, respectively. For each user $u \in \mathcal{U}$, we denote their chronologically ordered interaction history by $\mathbf{s}_u = (i_1, \dots, i_t)$, where $i_j \in \mathcal{I}$ $\forall \, j$. For each item $i \in \mathcal{I}$, we assume access to metadata: a textual description $d_i$, and (if available) a rating $\rho_{u,i}$ and review $v_{u,i}$ provided by user-$u$. Consequently, an interaction $(u,i)$ can be represented as the tuple $(i, d_i, \rho_{u,i}, v_{u,i}).$ Given the above, for each user $u \in \mathcal{U}$, the task is to predict the next item $i_{t+1}$ from a fixed candidate set $\mathcal{C}_u = \{ c_1, \dots, c_{10} \}$, which contains the ground-truth item and nine other non-interacted items.

\textit{Candidate set, $\mathcal{C}_u,$ generation.}
Following prior work \cite{kim2024large}, we adopt SASRec \cite{kang2018self} as the backbone sequential model for generating $\mathcal{C}_u$.
Given $\mathbf{s}_u$, SASRec produces a contextual state $\mathbf{z}^u_t$ and is trained with the standard next-item prediction objective given as: $\max_{\Theta} \sum_{u \in \mathcal{U}} \sum_{k=1}^{|\mathbf{s}_u|-1}
\log p\!\left(i_{k+1}\mid i_1{:}i_k;\,\Theta\right),$

where $p(\cdot)$ is the model-implied probability. 
At evaluation, $\mathcal{C}_u$ contains the ground-truth item $i_{t+1}$ and a subset of non-interacted items.

\textit{Metrics.} To evaluate a model's performance, we use the standard top-$1$ ranking metric, Hit Ratio at 1 (HR@1), given as: $\mathrm{HR@1}
= \frac{1}{|\mathcal{U}_{\text{test}}|}
\sum_{u\in\mathcal{U}_{\text{test}}}
\mathbb{I}\!\left\{ \pi_u(i_u^\ast) = 1 \right\},$

where $\mathcal{U}_{\mathrm{test}}$ is the set of users present in the test data, $i_u^\ast \in \mathcal{C}_u$ is the ground-truth next item for user $u$. For the sake of convenience, we keep all the notation used in this work in Table~\ref{tab:notations}.

\section{Proposed Model}
\label{sec:proposed-model}
In this section we describe the proposed model in detail. Our proposed model contains two phases.  
\subsection{Phase I: Generation of Thought Space}
\begin{figure}[h]
  \centering
  \includegraphics[width=1.0\textwidth]{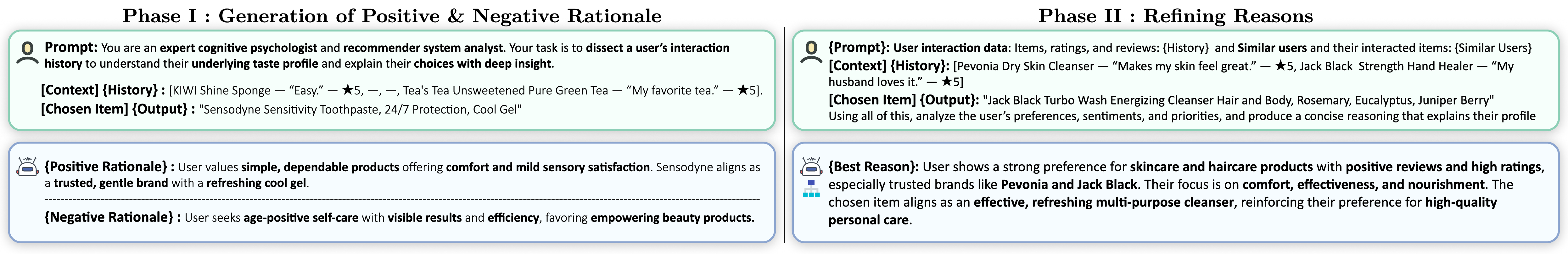}
  % \includesvg[width=0.8\linewidth]{figures/prompting-framework.svg}
  \small
  \caption{Two-stage prompting framework. In Phase 1, the LLM 
  (\includegraphics[height=1.2em]{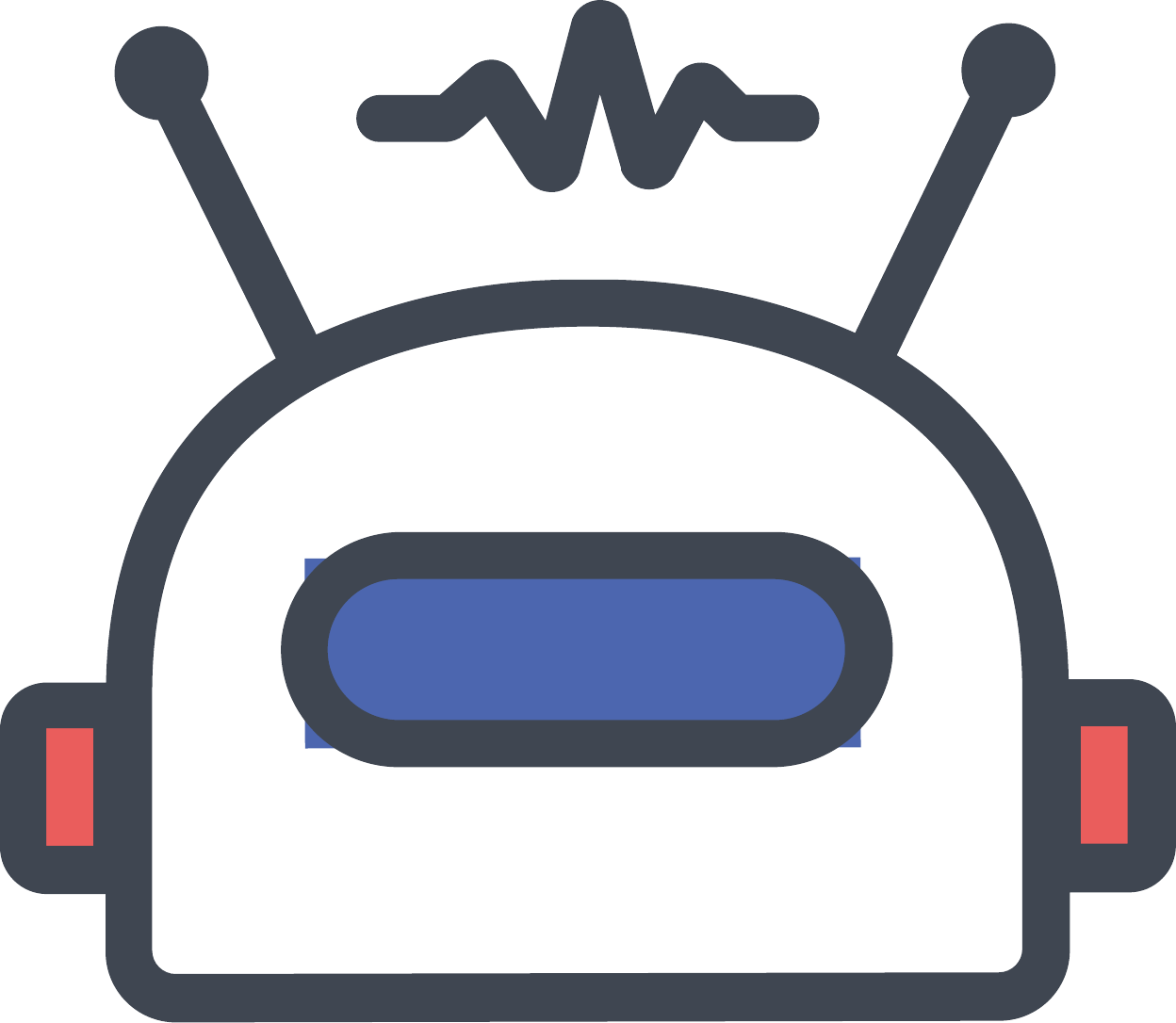}) generates a rationale from the user’s history and choice. In Phase 2, reasoning is refined by human-like oversight 
  (\includegraphics[height=1.2em]{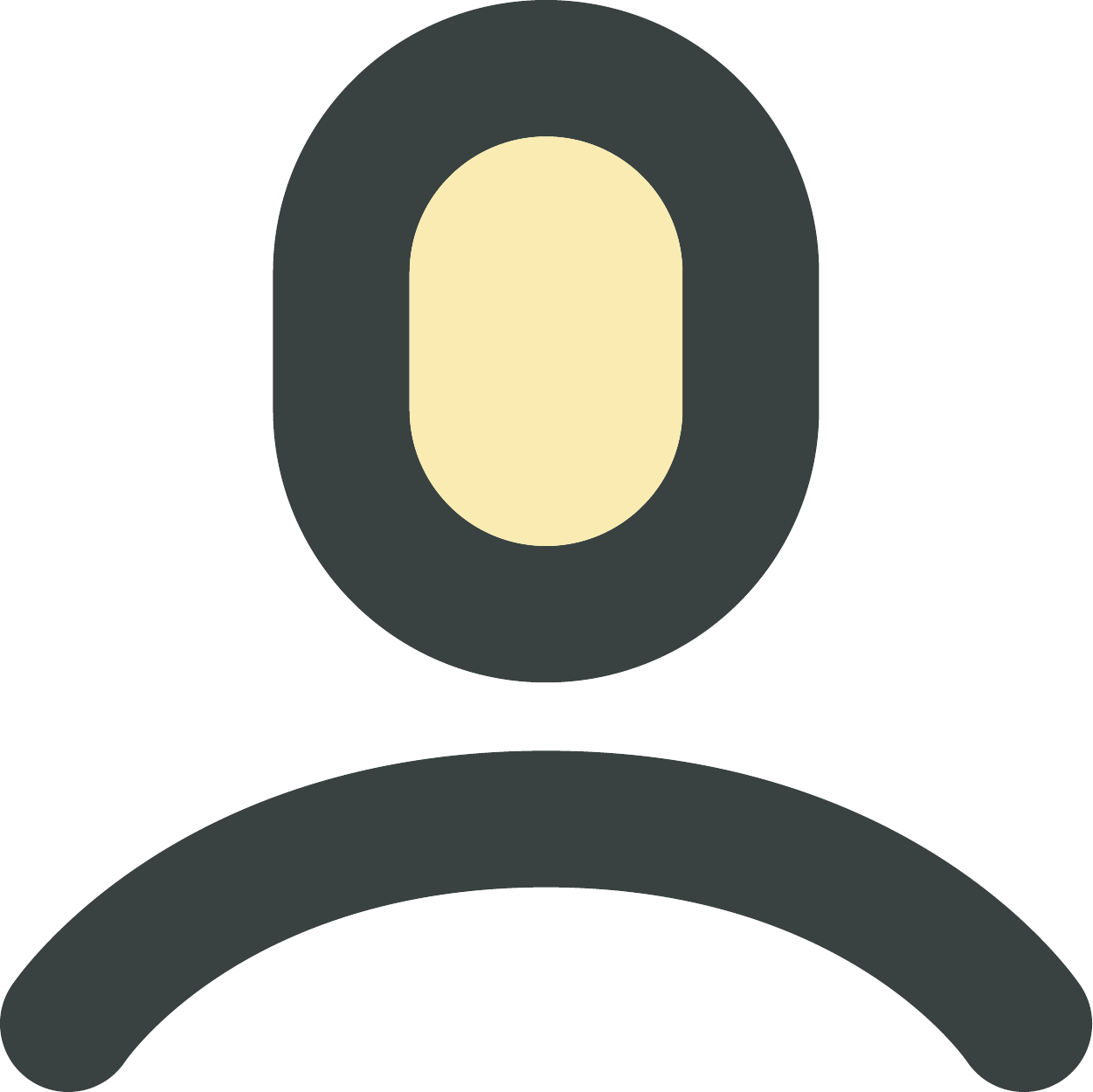}) combined with Tree-of-Thought exploration 
  (\includegraphics[height=1.2em]{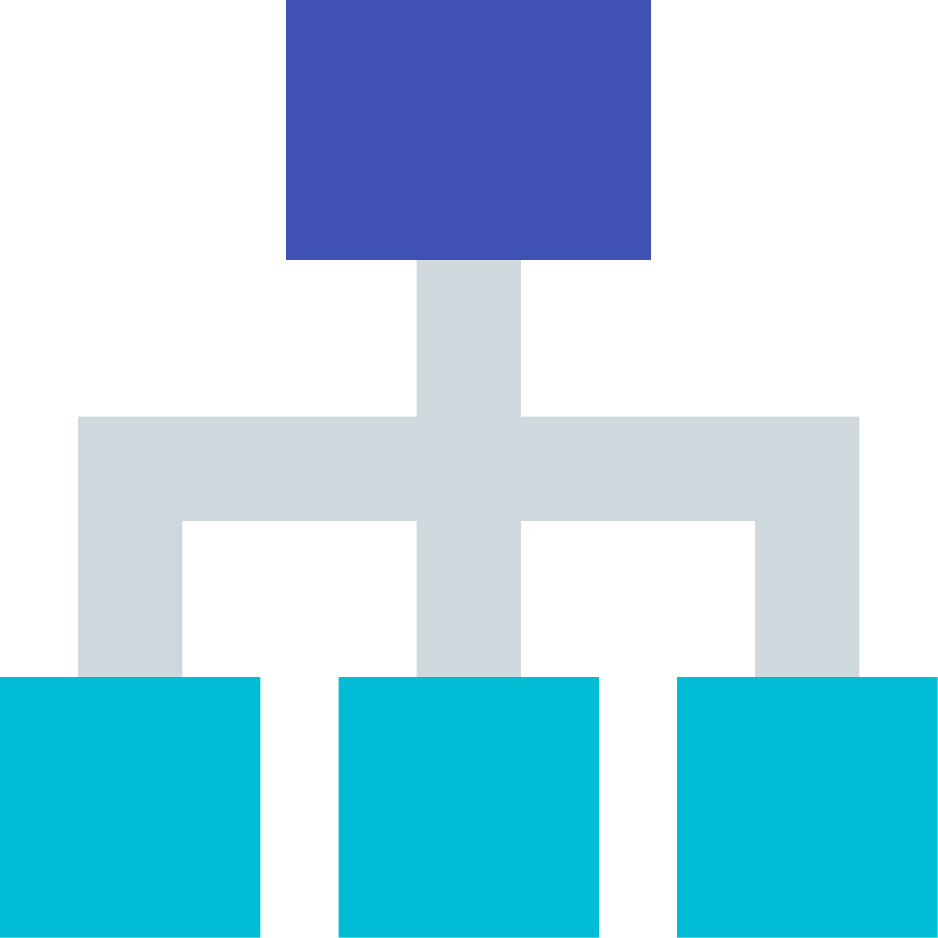}) to 
  derive the best reason. 
  [Phase 1: LLM rationale generation; Phase 2: LLM + ToT reasoning]}
  \label{fig:two_stage_prompting}
\end{figure}

We consider input data comprising user-item interactions, along with the reviews and ratings provided by users for the items they have engaged with. The output data consists of the ground-truth answers, defined as the next item a user interacts with following their interaction history. Our first objective is to generate thinking tokens (reasoning tokens or rationale tokens), denoted by $R,$ for these ground-truth outputs, conditioned on the interactions, reviews, and ratings. To this end, we prompt an SLM, specifically Phi-4 (4B), to derive the thinking tokens from the given input–output pairs. These tokens are referred to as \emph{positive rationales} $R_u^+$, as they correspond to user behaviors reflected in the ground-truth outputs based on the observed inputs. An example prompt is shown in Figure~\ref{fig:two_stage_prompting}.
\begin{figure}
    \centering
    \includegraphics[width=0.9\textwidth]{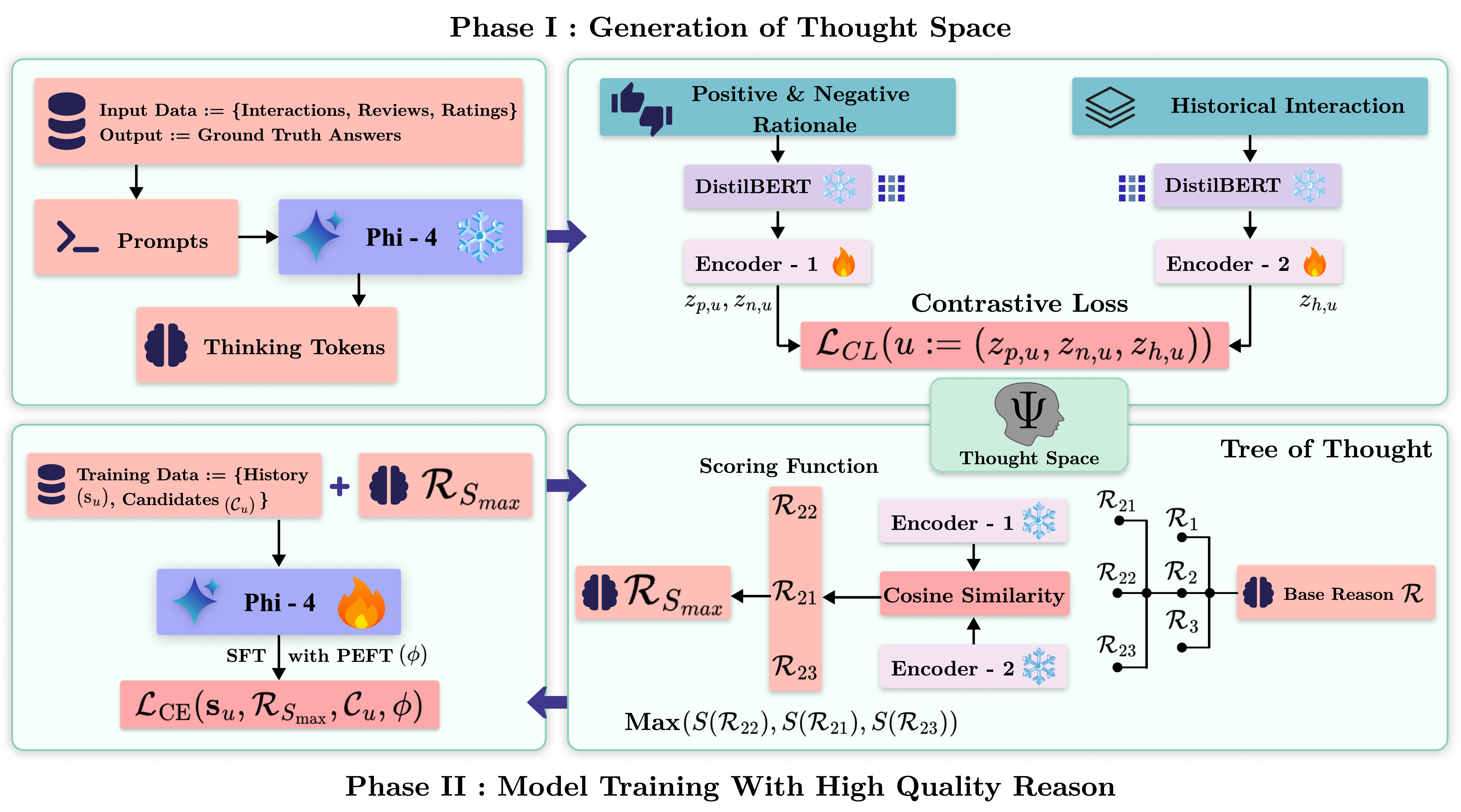}
    \caption{The architecture diagram. 
    Components marked with 
    \includegraphics[width=1em]{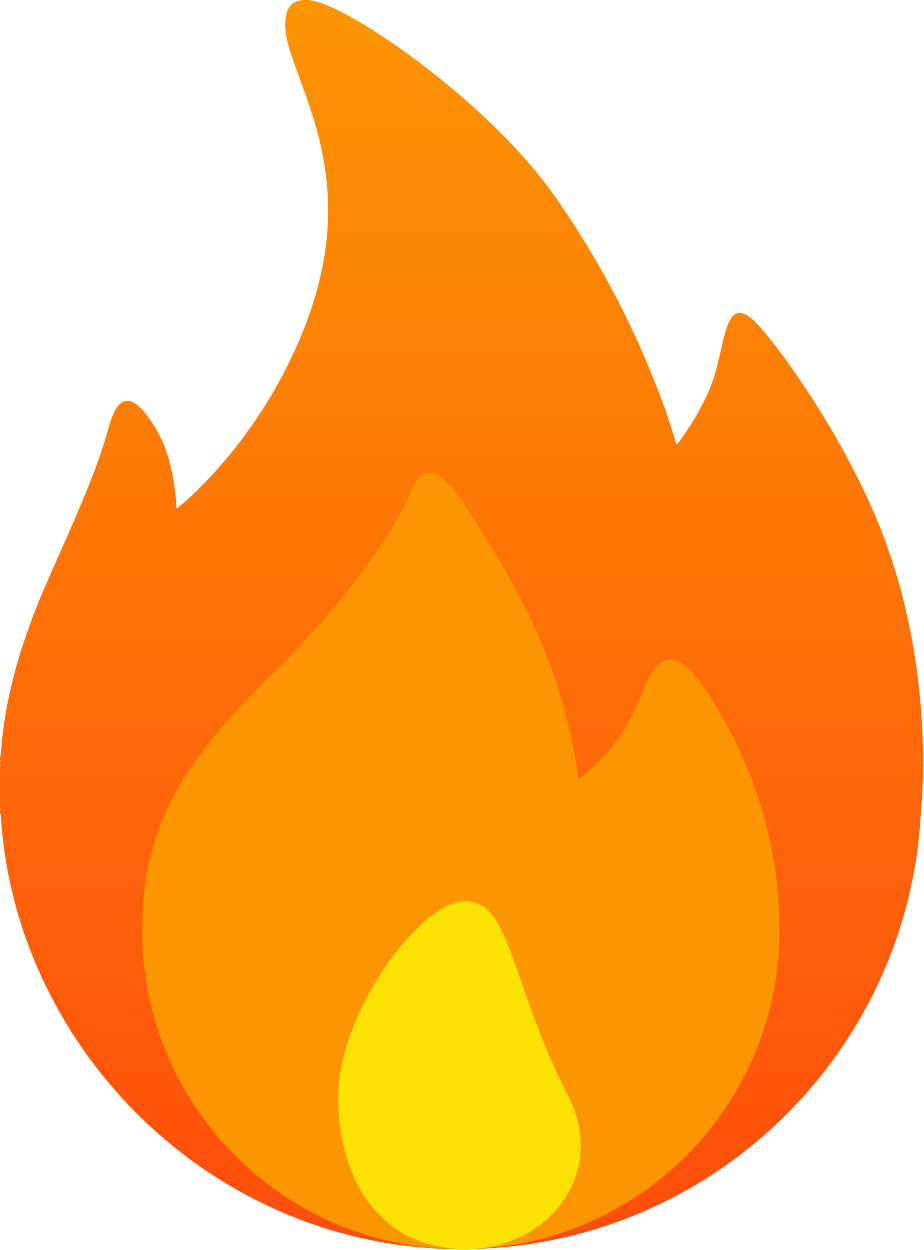}
    are \textbf{trainable}, whereas those marked with 
    \includegraphics[width=1em]{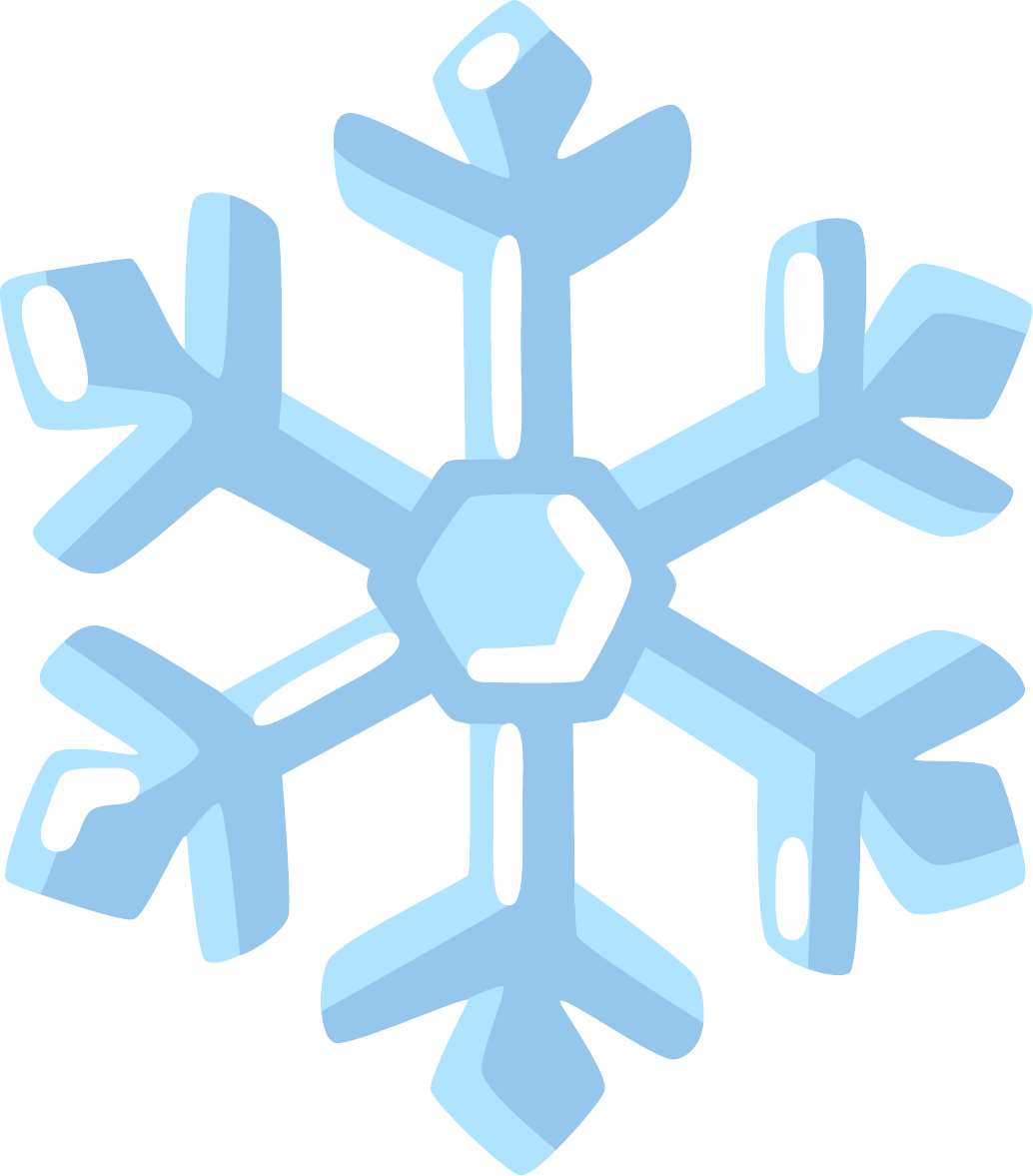}
    are \textbf{frozen} and only used during inference.}
    \label{fig:proposed-architecture}
\end{figure}

We now construct the \textbf{Thought Space} from the previously generated thinking tokens as follows. First, we generate negative rationales $R_v^-$ for all users. For a given user $u$, the \emph{negative rationales} $\{R_{v_j}^-\}_{j=1}^N$ are defined as the positive rationales of other users $v_j \neq u$ in the mini-batch $N$, where for each positive rationale we randomly sample 10 negative rationales to maintain a 1:10 ratio.
In parallel, we define the \emph{behavioral text} $b_u = (\mathbf{s}_u, i_{t+1})$, which concatenates the user’s interaction history $\mathbf{s}_u$ with the ground-truth item $i_{t+1}$. 

Now, we initialize two encoders, namely (i) a \textbf{\emph{Rationale Encoder} $E_1$} mapping rationales to embeddings $\mathbf{z}_r = E_1(R)$, and  (ii) a \textbf{\emph{History Encoder} $E_2$} mapping behavioral text to embeddings $\mathbf{z}_h = E_2(b_u)$. Here, both these encoders are fine tuned simultaneously with a contrastive learning objective. Let $\mathbf{z}_{p,u} = E_1(R_u^+)$, $\mathbf{z}_{n,u} = E_1(R_{v_j}^-)$, and $\mathbf{z}_{h,u} = E_2(b_u)$ denote the embeddings of the positive rationale, negative rationales, and history for user $u$, respectively. We train the encoders (E1 and E2 in Figure~\ref{fig:proposed-architecture}) with a contrastive objective that aligns $\mathbf{z}_{p,u}$ with $\mathbf{z}_{h,u}$ while separating $\mathbf{z}_{n,u}$. Formally, we optimize an InfoNCE loss:
\begin{equation}
\small
\mathcal{L}_{\text{CL}}(u)
= - \log \frac{\exp\!\big(\operatorname{sim}(\mathbf{z}_{p,u}, \mathbf{z}_{h,u})/\tau\big)}
{\exp\!\big(\operatorname{sim}(\mathbf{z}_{p,u}, \mathbf{z}_{h,u})/\tau\big) + \sum_{j=1}^{M} \exp\!\big(\operatorname{sim}(\mathbf{z}_{n,u}, \mathbf{z}_{h,u})/\tau\big)},
\label{eq:contrastive-loss-phase-1}
\end{equation}
where $\operatorname{sim}(\cdot, \cdot)$ is the cosine similarity and $\tau>0$ is temperature. This contrastive training yields a \emph{Thought Space} $\subseteq \mathbb{R}^d$, where rationales consistent with a user’s behavior are aligned with user's historical interactions, and inconsistent rationales are pushed apart.

\begin{figure}[ht!]
    \centering
    \includegraphics[width=0.9\textwidth]{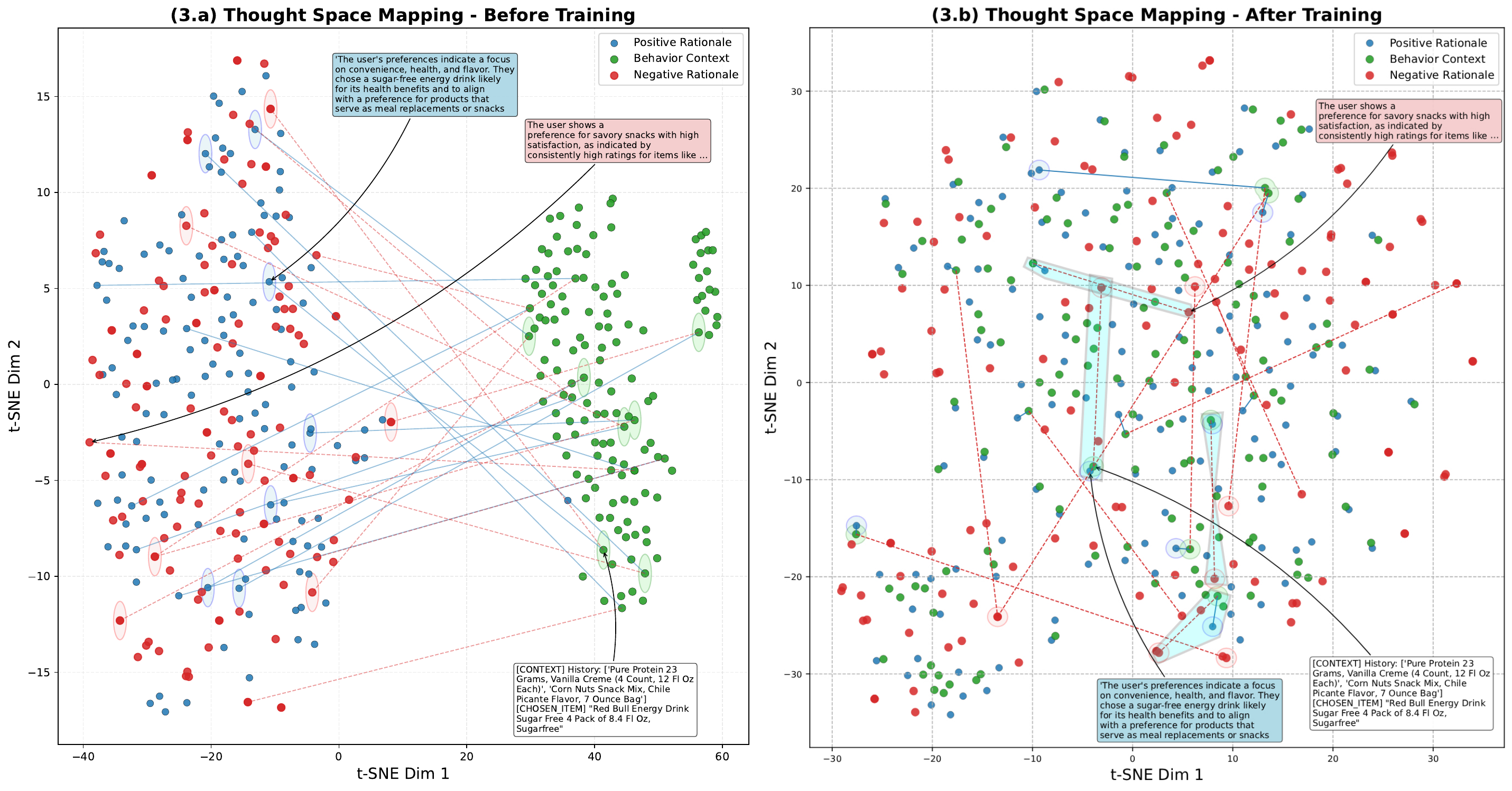}
    \small
    \caption{\textbf{Thought Space before vs. after contrastive alignment}. (a) Pre-training: behavioral embeddings (green) and rationale embeddings with positives (blue) and negatives (red)are misaligned, with DistilBERT-initialized behavioral and rationale texts occupying different regions. (b) Post-training: after optimizing Eq. (1), positives cluster near their behavioral anchors (blue edges shorten), while negatives are repelled (red edges lengthen).}
    \label{fig:ThoughtSpaceAlignment}
\end{figure}

Figure~\ref{fig:ThoughtSpaceAlignment} shows t\textsc{-SNE} projections of (i) behavioral embeddings $\mathbf{z}_{h,u}$ (green nodes), (ii) positive rationale embeddings $\mathbf{z}_{p,u}$ (blue nodes), and (iii) negative rationale embeddings $\mathbf{z}_{n,u}$ (red nodes), \emph{before} and \emph{after} contrastive training. Prior to training, the spaces are poorly aligned although $E_1$ and $E_2$ are
initialized from DistilBERT, behavioral text and rationale text occupy noticeably different regions. After optimizing Equation \eqref{eq:contrastive-loss-phase-1}, positives concentrate near their corresponding behavioral anchors, while negatives are repelled: blue nodes move closer to the green nodes, and the blue solid edges (nearest-neighbor links from each $\mathbf{z}_{h,u}$ to its $\mathbf{z}_{p,u}$) become visibly shorter; conversely, red nodes drift away and the red dotted edges (links to $\mathbf{z}_{n,u}$) lengthen. This geometric separation visualizes the intended alignment in the Thought Space and anticipates its utility in downstream selection and ranking.
%\vspace{-0.5cm}
\begin{table}
\centering
\small
\caption{Statistics of the Amazon datasets used in our experiments.}
\label{tab:datasets}
\begin{tabular}{|l|c|c|c|c|}
\hline
\textbf{Dataset} & \textbf{\#Users} & \textbf{\#Items} & \textbf{Avg. Seq. Length} & \textbf{\# of Interactions}\\
\hline
Luxury Beauty & 11{,}690 & 6{,}534 & 4.15 & 71898\\
\hline
Prime Pantry  & 15{,}611 & 7{,}841 & 7.64 & 150402 \\
\hline
Video Games   & 64{,}073 & 33{,}614 & 6.88 & 568508 \\
\hline
\end{tabular}
\end{table}
\subsection{Phase II. Model training with high quality reasons}
In this section, we propose a Tree-of-Thought (ToT) approach for obtaining the best reasons from a pool of candidate base reasons as illustrated in Figure~\ref{fig:proposed-architecture} bottom right block, followed by SFT to get the final recommendation. 

\textbf{Tree-of-Thought Refinement.}
We first generate a base reason $\mathcal{R}$ from a simple prompt as seen in Phase~II of Figure~\ref{fig:two_stage_prompting} (conditioned on user history $\mathbf{s}_u$ and items of similar users). The similar users are computed using cosine similarity on the SASRec embeddings obtained in Phase~I. We then refine the base reason using a Tree-of-Thought (ToT) approach. Specifically, the small language model (SLM; Phi-4, 4B) expands $\mathcal{R}$ into $n$ first-level refinements $\{\mathcal{R}_1,\dots,\mathcal{R}_n\}$. Each $\mathcal{R}_i$ is further expanded into $m$ second-level refinements $\{\mathcal{R}_{i,1}, \mathcal{R}_{i,2}, \dots, \mathcal{R}_{i,m}\}$. In our setup, the tree depth is two, but it can be generalized to arbitrary depths.

Each leaf rationale $\mathcal{R}_{i,j}$ is embedded via the rationale encoder $E_1$, producing $\mathbf{z}_{r} = E_1(\mathcal{R}_{i,j})$. The corresponding user behavior context $b_u=(\mathbf{s}_u, i_{t+1})$ is mapped by the history encoder $E_2$ to $\mathbf{z}_h = E_2(b_u)$. A scoring function $S$ based on cosine similarity evaluates the agreement between rationale and behavior embeddings:
$
S(\mathcal{R}_{i,j}) = \operatorname{sim}\!\big(E_1(\mathcal{R}_{i,j}), E_2(b_u)\big).
$

We denote by $S_{\max}$ the \emph{highest agreement score} among all candidate rationales: $S_{\max} = \max_{\mathcal{R}_{i,j}} S(\mathcal{R}_{i,j}).$

Finally, we define $\mathcal{R}_{S_{\max}}$ as the \emph{best refined rationale}, i.e., the one achieving this maximum score:
\begin{equation}
\small
\label{eq:best-rationale}
\mathcal{R}_{S_{\max}} = \arg\max_{\mathcal{R}_{i,j}} \; S(\mathcal{R}_{i,j}).
\end{equation}

\textbf{Supervised Fine-Tuning (SFT).}
After selecting the best reason $\mathcal{R}_{S_{\max}}$ from the ToT stage in phase-II, we incorporate it into an SFT stage. For each user $u$, the inputs consist of their historical sequence $\mathbf{s}_u$, the fixed 10-item candidate set $\mathcal{C}_u$, and the best reason $\mathcal{R}_{S_{\max}}$ (with agreement score $S_{\max}$). In this stage, we train only the parameters of the SLM under a Parameter-Efficient Fine-Tuning (PEFT) configuration (LoRA), while the base encoders $E_1$ and $E_2$ remain frozen. The SLM with PEFT parameters $\phi$ produces a scalar logit for each candidate by conditioning on the user history and the selected reason:$ 
\, \ell_i = f_{\phi}\!\big(\,\mathbf{s}_u,\, \mathcal{R}_{S_{\max}},\, c_i\,\big),$ 
for $c_i \in \mathcal{C}_u,$
where $f_\phi$ denotes the SLM scoring head (only $\phi$ is trainable).
These logits, $(\ell_i ),$ are normalized with a softmax to obtain a categorical distribution over candidates. The training objective is the cross-entropy loss that maximizes the likelihood of the ground-truth next item $i^u_{t+1}$. The model is trained to identify the ground-truth item $i^u_{t+1}$ among the candidate set by minimizing the cross-entropy objective:
\begin{equation}
\small
\mathcal{L}_{\mathrm{CE}}(u) 
\;=\; 
- \log\, p_{\phi}\!\big(i^u_{t+1} \mid \mathbf{s}_u, \mathcal{R}_{S_{\max}}, \mathcal{C}_u\big).
\label{eq:ce-sft}
\end{equation}
\noindent
\textit{Notes.} (i) The selection of $\mathcal{R}_{S_{\max}}$ uses frozen $E_1,E_2$; no gradients back propagates to Phase~I. 
(ii) The optimization follows Eq.~\eqref{eq:ce-sft}.

This fine-tuning process ensures that the final recommender is conditioned not only on the interaction history but also on the rationale most consistent with the user’s behavior in the learned \emph{Thought Space}. As a result, the model captures both the sequential dynamics of user interactions and the semantic reasoning signals learned in Phase~I.

\section{Experiments}
\label{sec:experiments}
\subsection{Datasets}
\label{subsec:datasets}
We evaluate on three Amazon Product Review datasets spanning distinct domains and sparsity regimes: \emph{Luxury Beauty}, \emph{Prime Pantry}, and \emph{Video Games}, with statistics summarized in Table~\ref{tab:datasets}. We follow standard preprocessing: (i) de-duplicate interactions per user–item; (ii) chronologically sort interactions per user; (iii) filter users/items with fewer than 5 interactions; and (iv) tokenize item metadata (titles/descriptions) with the same vocabulary as encoders.
%\textit{Train/Val/Test Split.}
For each user $u$, we adopt a standard time-aware split: the last interaction is used as the test set, and the remaining interactions for training set. Sequences are truncated/padded to a maximum length~$50$.
%\textit{Candidate set protocol.}
At evaluation we rank the ground-truth next item among a fixed candidate set $\mathcal{C}_u$ of size 10 (1 ground truth $+$ 9 non-interacted items). Unless otherwise noted, negatives are sampled uniformly from the item universe excluding items in $\mathbf{s}_u$.
\subsection{Baselines}
\label{subsec:baselines}
We compare against strong CF based sequential and LLM/SLM-augmented recommenders:
\begin{itemize}
    \item \textbf{SASRec}~\cite{kang2018self}: Transformer-based sequential CF.
    \item \textbf{CTRL}~\cite{li2023ctrl}: A cross-modal framework that aligns collaborative signals from tabular CTR data with semantic signals from pre-trained language models.
    \item \textbf{MoRec}~\cite{yuan2023go}: A modality aware recommendation framework.
    \item \textbf{DuoRec} \cite{qiu2022contrastive}: A sequential recommender that mitigates representation degeneration via contrastive regularization.
    \item \textbf{EC4SRec}~\cite{wang2022explanation}: A sequential recommender that uses explanation-guided augmentations to generate semantically faithful positives and negatives, for supervised and self-supervised CL to improve sequence representation.
    \item \textbf{ALLMRec} \cite{kim2024large}: An LLM–CF hybrid that injects collaborative knowledge from a pre-trained state-of-the-art CF model into a LLM.
    \item \textbf{ALLMRec+Reason}: Variant of ALLMRec with reasons augmented by an SLM (Phi-4, 4B)
    \item \textbf{SLM+SFT}: SLM (Phi-4, 4B), SFT with interaction history and ground truth but without reasoning supervision.
\end{itemize}

\subsection{Implementation and Settings}
\label{subsec:impl}
Phase I (contrastive learning) uses batch size $B=32$, AdamW optimizer (lr=$2\!\times\!10^{-5}$ for encoders, $1\!\times\!10^{-3}$ for $f_\phi$), and linear warmup (10\%) followed by cosine decay. Phase II (SFT) uses batch size $B=1$ for training, $B=2$ at inference, and LoRA (rank $r=8$, $\alpha=16$, dropout=0.05) on the SLM scoring head. The learning rate is $2\!\times\!10^{-4}$ with AdamW. Maximum token length is 512. Early stopping is applied on validation HR@1. All experiments run on dual NVIDIA RTX~3090 GPUs (24GB each).

\begin{table}[h]
\centering
\caption{Overall performance comparison (HR@1) across datasets. 
The best result per dataset is in \textbf{bold} and the second best is underlined.}
\label{tab:overall-results}
\begin{tabular}{|l|c|c|c|}
\hline
\textbf{Model} & \textbf{Luxury Beauty} & \textbf{Prime Pantry} & \textbf{Video Games} \\
\hline
SASRec & 0.5612 & 0.3702 & 0.6117 \\
\hline
CTRL & 0.3154 & 0.2660 & 0.4003 \\
\hline
MoRec & 0.3029 & 0.1300 & --\footnotemark \\
\hline
DuoRec & 0.5412 & 0.3519 & 0.5139 \\
\hline
EC4SRec & 0.5781 & 0.3995 & 0.5891 \\
\hline
ALLMRec & 0.6014 & 0.6237 & 0.6990 \\
\hline
ALLMRec+Reason & \underline{0.7309} & \underline{0.7426} & \underline{0.7781} \\
\hline
SLM+SFT & 0.6100 & 0.5700 & 0.6757 \\
\hline
\textbf{PULSE} & \textbf{0.9339} & \textbf{0.8373} & \textbf{0.9170} \\
\hline
\end{tabular}
\end{table}
\footnotetext{Unavailable due to computational constraints.}

\subsection{Overall Performance}
We compare our proposed method PULSE with the baselines mentioned in Section~\ref{subsec:baselines} on the datasets given in Section~\ref{subsec:datasets}. As seen in Table~\ref{tab:overall-results}, our method consistently outperforms all baselines by a substantial margin, improving HR@1 by up to 9--20\% over strong reasoning-based models such as ALLMRec+Reason which is simply ALLMRec model complemented with reasons or user profiles. Notably, despite relying on an SLM backbone, our approach surpasses billion-scale LLM-based baselines, validating the effectiveness of leveraging rationales as learning signals in the 
\emph{Thought Space}.

\subsection{Ablation Studies}
\label{subsec:ablations}
\textbf{Ablation A:} Which embedding space should score rationales? To isolate the contribution of our \emph{Thought Space}, we hold the entire pipeline fixed, including SASRec candidate generation, the ToT expansion producing a pool of candidate reasons $\{\mathcal{R}_{i,j}\}$, and the downstream SFT recipe, and vary only the \emph{scoring space} used to select the single rationale per user. We compare:
\begin{enumerate}
    \item \textbf{Thought Space (ours).} Agreement is measured with Phase~I encoders: 
    $S_{\text{TS}}(R)=\operatorname{sim}\!\big(E_1(R),E_2(b_u)\big)$.
    
    \item \textbf{Vanilla DistilBERT.} Replace $E_1/E_2$ with a frozen DistilBERT $D(\cdot)$ on raw text; score by cosine: 
    $S_{\text{DB}}(R)=\operatorname{sim}\!\big(D(R),D(b_u)\big)$.
    
    \item \textbf{SBERT.} Use a frozen Sentence-BERT encoder (all-MiniLM-L6-v2) $S(\cdot)$; score by cosine: 
    $S_{\text{SB}}(R)=\operatorname{sim}\!\big(S(R),S(b_u)\big)$.
\end{enumerate}
For each scoring method, we select the best reason, $\mathcal{R}_{S_{\max}},$ as per Eq.~\eqref{eq:best-rationale},
%\mathcal{R}_{S_{\max}} \;=\; \arg\max_{\mathcal{R}} S(\mathcal{R}),
to construct a \emph{distinct} SFT dataset with those chosen rationales, and train the same PEFT head $f_\phi$ under identical hyperparameters. Final recommendation metrics (HR@1) are reported in Table~\ref{tab:ablation_spaces}. This isolates whether \emph{behavior--reason alignment learned via contrastive training} (Thought Space) yields gains beyond generic semantic similarity (DistilBERT/SBERT).
\begin{table}
\centering
\small
\caption{Rationale scoring spaces (HR@1).}
\label{tab:ablation_spaces}
\renewcommand{\arraystretch}{1.2}
\begin{tabular}{|l|c|c|c|}
\hline
\textbf{Scoring space} & \textbf{Luxury Beauty} & \textbf{Prime Pantry} & \textbf{Video Games} \\
\hline
DistilBERT cosine  & 0.7920 & 0.6284 & \underline{0.8896} \\
\hline
SBERT cosine       & \underline{0.8636} & \underline{0.7182} & 0.80459 \\
\hline
\textbf{Thought Space (ours)} & \textbf{0.9339} & \textbf{0.8373} & \textbf{0.9170} \\
\hline
\end{tabular}
\end{table}

\textbf{Ablation B:} Which SFT configuration helps most? We now fix the rationale scoring to the best option from Ablation~A and compare \emph{how} supervised fine-tuning (SFT) uses reasoning. We evaluate four settings:
\begin{enumerate}
\item \textbf{SFT w/o reasons} (\textit{History + Candidates only}). Establishes a text-free baseline.
\item \textbf{SFT + base reason} (\textit{History + Candidates + single prompt reason}). Tests the value of any explanation text.
\item \textbf{SFT + ToT (Log-Likelihood (\texttt{LL}) score).} Generate multiple rationales via ToT; select by \texttt{LL}, under the SLM; SFT on the selected rationale. Measures benefit of ToT while keeping a standard generative selector.
\item \textbf{SFT + ToT (Thought Space)} \textit{(ours).} Same ToT pool, but select $\mathcal{R}_{S_{\max}}$ by agreement in Thought Space (Eq.~\ref{eq:best-rationale}); SFT on that rationale. Quantifies the added value of behavior-aligned selection over likelihood.
\end{enumerate}

All runs share the same PEFT setup, token limits, optimizer, and SASRec candidates.
Results in Table~\ref{tab:ablation_sft} disentangle three effects:
(i) adding reasons at all \((1\!\rightarrow\!2)\),
(ii) using structured multi-step generation \((2\!\rightarrow\!3)\), and
(iii) replacing likelihood with Thought Space \((3\!\rightarrow\!4)\).
\vspace{-0.5cm}
\begin{table}
\centering
\small
\caption{SFT variants with/without reasons and different rationale selectors.}
\label{tab:ablation_sft}
\renewcommand\arraystretch{1.2} % adds a bit of row spacing
\begin{tabular}{|l|l|c|c|c|}
\hline
\textbf{No.} & \textbf{SFT configuration} 
& \shortstack{\textbf{Luxury} \\ \textbf{Beauty}} 
& \shortstack{\textbf{Prime} \\ \textbf{Pantry}} 
& \shortstack{\textbf{Video} \\ \textbf{Games}} \\
\hline
1. & History + Candidates (no reasons) & 0.6111 & 0.5748 & 0.6757 \\
\hline
2. & 1 + Base prompt reason & 0.7186 & \underline{0.7718} & \underline{0.8883} \\
\hline
3 & 2 + ToT, selected by \texttt{LL} & \underline{0.8559} & 0.7548 & 0.8647 \\
\hline
4 & 2 \textbf{+ ToT, selected by Thought Space (ours)} & \textbf{0.9339} & \textbf{0.8373} & \textbf{0.9170} \\
\hline
\end{tabular}
\end{table}
\vspace{-0.5cm}
Ablation~A tests \emph{where} to measure agreement (generic semantic spaces vs. our contrastively learned Thought Space). Ablation~B tests \emph{how} to use reasons in SFT (none vs. single reason vs. ToT with likelihood vs. ToT with Thought-Space selection), cleanly attributing gains to \emph{reason presence}, \emph{structured generation}, and \emph{behavior-aligned selection}.

\subsection{Cross-Domain Recommendation}
\label{sec:cross-domain}
To test robustness under distribution shift, we train on \emph{Luxury Beauty} and evaluate directly on \emph{Video Games}. This setting is difficult since user interests and item semantics differ sharply. Results in Table~\ref{tab:cross-domain} show that SASRec (HR@1=0.103) and MoRec (0.070) collapse under shift, confirming that collaborative priors and modality-only signals fail to generalize. SFT improves: adding no reasons yields 0.4075, while appending base rationales raises HR@1 to 0.558. Using Tree-of-Thought rationales with \texttt{LL} selection underperforms at 0.546, indicating that generative confidence is unstable under domain transfer.  

Our proposed \textbf{PULSE}, which selects rationales via \emph{Thought Space} alignment, achieves the best result (HR@1=0.624), with gains of \textbf{11.8\%} over base reasons, \textbf{14.3\%} over log-likelihood, and \textbf{53.1\%} over SFT w/o reasons. Relative to SASRec, PULSE is $6.1\times$ stronger and nearly $9\times$ better than MoRec. These results show that reasoning-level contrastive alignment yields domain-robust signals by emphasizing \emph{why} users act, not just \emph{what} they consumed in the source domain.
\begin{table}
\centering
\begin{minipage}{0.49\textwidth}
\centering
\small
\caption{Cross-domain (Train: Luxury~Beauty $\rightarrow$ Test: Video~Games), HR@1.}
\label{tab:cross-domain}
\begin{tabular}{|l|c|}
\hline
\textbf{Model} & \textbf{HR@1} \\
\hline
SASRec & 0.1030 \\
\hline
MoRec & 0.0700 \\
\hline
CTRL & 0.0673 \\
\hline
SFT (w/o reasons) & 0.4075 \\
\hline
SFT (base reasons) & 0.5580 \\
\hline
SFT (ToT +LL) & 0.5460 \\
\hline
\textbf{PULSE} & \textbf{0.6240} \\
\hline
\end{tabular}
\end{minipage}
\hfill
\begin{minipage}{0.49\textwidth}
\centering
\small
\caption{HotpotQA development set: F1 and Exact Match (EM). Thought Space is used only to select the rationale; no other components are changed.}
\label{tab:hotpotqa}
\begin{tabular}{|l|c|c|}
\hline
\textbf{Method} & \textbf{F1} & \textbf{EM} \\
\hline
Standard prompting & 0.5607 & 0.4553 \\
\hline
CoT (chain-of-thought) & 0.5623 & 0.4553 \\
\hline
Act (w/o reasoning)  & 0.5318 & 0.4333 \\
\hline
ReAct + CoT & 0.5667 & 0.4283 \\
\hline
\shortstack{\textbf{ReAct +} \\ \textbf{Thought Space (ours)}} & \textbf{0.6060} & \textbf{0.4951} \\
\hline
\end{tabular}
\end{minipage}
\end{table}
\subsection{Thought Space Improves Question Answering on HotpotQA}
\label{sec:hotpotqa}

To test whether \emph{Thought Space} captures transferable reasoning, we evaluate on HotpotQA (distractor setting), a multi-hop QA benchmark. We compare against Standard Prompting, CoT, Act, and ReAct+CoT under identical prompts and backbones. Our method differs only in rationale selection: multiple candidate rationales are generated (Gemma-3, 27B), scored with Thought Space encoders, and the highest-agreement rationale $\mathcal{R}_{S_{\max}}$ is retained. This produces training triples $(q, \mathcal{R}_{S_{\max}}, a)$, used to fine-tune an SLM (Phi-4, 4B) under PEFT.

As shown in Table~\ref{tab:hotpotqa}, incorporating Thought Space improves both F1 and EM. Relative to Standard/CoT (0.560/0.450), we achieve \textbf{+8.2\%} F1 and \textbf{+10.0\%} EM. Against ReAct+CoT, the F1 gain is the same but EM rises by \textbf{+17.9\%}. Compared to Act, Thought Space yields the largest lift: \textbf{+14.3\%} F1 and \textbf{+15.1\%} EM. The best variant, \emph{ReAct with Thought Space}, demonstrates that reasoning-level contrastive alignment learned in recommendation transfers effectively to QA, underscoring the generality of our approach.

\vspace{-0.1cm}
\section{Conclusion}
\label{sec:conclusion}
We presented PULSE, a reasoning-augmented recommender that aligns user histories and rationales in a shared Thought Space. Unlike prior work that treats rationales as auxiliary text or selects them by likelihood, PULSE uses contrastive learning to ground rationales in behavioral context and then fine-tunes a small language model on the most consistent explanation. Across three Amazon domains, this approach achieves state-of-the-art sequential recommendation performance, including strong gains in cross-domain transfer, where Thought Space proves more robust than semantic similarity baselines.

Our results highlight that reasoning as a supervised signal improves recommendation and general reasoning, with PULSE outperforming agentic baselines on HotpotQA and enabling compact models to surpass billion-parameter LLMs. These findings show that compact models, when aligned with reasoning signals, can outperform billion-parameter LLMs in accuracy, efficiency, and generalization. Future work will extend PULSE beyond text to multimodal domains, paving the way for scalable, interpretable reasoning-aware recommenders.

\bibliographystyle{splncs04}
\bibliography{ecir-2025}

\end{document}